%% file: main.tex
\documentclass[letterpaper, 10 pt, conference]{ieeeconf}
\pdfoutput=1

\IEEEoverridecommandlockouts
\let\labelindent\relax

\makeatletter
\let\NAT@parse\undefined
\makeatother

\usepackage{amsmath,amssymb,amsfonts}
\usepackage{enumitem}
\usepackage{caption}
\usepackage{subcaption}
\usepackage{algorithmic}
\usepackage{array}
\usepackage{multirow}
\usepackage{graphicx}
\usepackage[dvipsnames]{xcolor}
\usepackage{hyperref}
\usepackage{textcomp}
\usepackage{bm}

\DeclareMathOperator*{\argmax}{arg\,max}
\DeclareMathOperator*{\const}{const}

\usepackage{float}

\newfloat{program}{thp}{lop}
\floatname{program}{Code Snippet}

\usepackage[newfloat,frozencache,cachedir=./mintedcache]{minted}

\usepackage[framemethod=latex]{mdframed}
\mdfdefinestyle{example}{
    innerleftmargin=0.1cm,innerrightmargin=0.1cm,roundcorner=10pt,
    backgroundcolor=lightgray!50,shadow=true,
    shadowsize=2pt,innermargin=2pt,outermargin=2pt,
    linewidth=0.8pt, linecolor=black,
    usetwoside=true,nobreak=true}

\usepackage[
protrusion=true,
activate={true,nocompatibility},
final,
tracking=true,
kerning=true,
spacing=true,
factor=1100]{microtype}%

\newcolumntype{P}[1]{>{\centering\arraybackslash}p{#1}}

\newcommand\dunderline[3][-.5pt]{{%
  \sbox0{#3}%
  \ooalign{\copy0\cr\rule[\dimexpr#1-#2\relax]{\wd0}{#2}}}}

\newif\ifarxiv
\arxivtrue

\ifarxiv
\newcommand{\citemore}[2]{%
    \ifx&#1&%
      \cite{#2}%
    \else%
      \ifx&#2&%
        \cite{#1}%
      \else%
        \cite{#1,#2}%
      \fi%
    \fi%
}%
\else
\newcommand{\citemore}[2]{%
  \ifx&#1&%
    \ignorespaces%
  \else%
    \cite{#1}%
  \fi%
}%
\fi

\begin{document}
\title{\LARGE \bf Visual Place Recognition: A Tutorial}
\author{
Stefan Schubert$^\dag$, Peer Neubert$^\ddag$, Sourav Garg$^{\dag\dag}$, Michael Milford$^{\ddag\ddag}$ and Tobias Fischer$^{\ddag\ddag}$
\thanks{$^\dag$ S.~S. is with the Chemnitz University of Technology, Germany.\newline Email: stefan.schubert@etit.tu-chemnitz.de}%
\thanks{$^\ddag$ P.~N. is with the University of Koblenz, Germany.\ifarxiv\newline Email: neubert@uni-koblenz.de\fi}%
\thanks{$^{\dag\dag}$ S.~G. is currently with the University of Adelaide, Australia (work done while at the Queensland University of Technology, Australia).\ifarxiv\newline Email: sourav.garg@adelaide.edu.au\fi}%
\thanks{$^{\ddag\ddag}$ M.~M.~and T.~F. are with the QUT Centre for Robotics, Queensland University of Technology, Australia.\ifarxiv\newline Email: \{michael.milford, tobias.fischer\}@qut.edu.au\fi}
\thanks{S.~S.~acknowledges support by the German Federal Ministry for Economic Affairs and Climate Action. S.~G., M.~M.~and T.~F.~acknowledge support by the QUT Centre for Robotics, funding from ARC Laureate Fellowship FL210100156 to M.~M., and a grant from Intel Labs to T.~F.~and M.~M.}%
\thanks{The authors would like to thank Dr Mark Zolotas, Dr Alejandro Fontan and Somayeh Hussaini for valuable insights on drafts of the paper.}
}

\maketitle
\pagestyle{empty}
\thispagestyle{empty} %

\input{1-Introduction.tex}
\input{2-History.tex}
\input{3-CategoriesUsecases.tex}
\input{4-Pipeline.tex}

\input{5-Evaluation.tex}
\input{6-Challenges.tex}

\input{7-Conclusions.tex}

\ifarxiv
\IEEEtriggeratref{94}
\else
\IEEEtriggeratref{35}
\fi
\bibliographystyle{IEEEtran}
\bibliography{references}

\end{document}

%% file: 1-Introduction.tex
\section{Introduction}
\label{sec:intro}
Localization is an essential capability for mobile robots, enabling them to build a comprehensive representation of their environment and interact with the environment effectively towards a goal.
A rapidly growing field of research in this area is Visual Place Recognition (VPR), which is the ability to recognize previously seen places in the world based solely on images.
The volume of published research on VPR has shown a significant and continuous growth over the years, from two papers with ``visual place recognition'' and seven papers with ``place recognition'' in the title in 2006, to 65 and 163 papers, respectively, in 2022\footnote{Source: Google Scholar with query \textit{allintitle: ``title''}}.
A number of survey and benchmarking papers have discussed the challenges, open questions, and achievements in the field of VPR \citemore{Schubert2021,Garg2021,Masone2021,Lowry2016}{zaffar2021vpr,Zhang2021}.
\begin{figure}[t]
    \centering
    \includegraphics{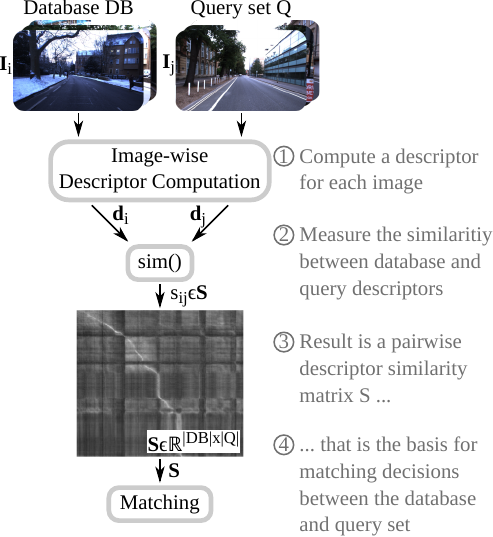}
    \caption{This figure illustrates the key steps and components of VPR as outlined in \autoref{sec:intro}. Historically, the matching decisions in step 4 were used for loop closure in Simultaneous Localization and Mapping (SLAM); \autoref{sec:history} provides an overview of the history and relevance of the VPR problem. While this figure illustrates a common use case where incoming imagery in the query set is compared to a database, \autoref{sec:taxonomy} distinguishes different VPR problem categories based on this pipeline and also relates them to VPR use cases. The details of each shown computational step will be discussed in \autoref{sec:vpr_problem}, followed by details on the evaluation of VPR pipelines in \autoref{sec:evaluation}.
    (Photos from \cite{ds_robotcar})}
    \label{fig:basic_pipeline}
\end{figure}

This present work is the first tutorial paper on visual place recognition.
It unifies the terminology of VPR and complements prior research in two important directions: 
\begin{enumerate}
 \item It provides a systematic introduction for newcomers to the field, covering topics such as the formulation of the VPR problem, a generic algorithmic pipeline, an evaluation methodology for VPR approaches, and the major challenges for VPR and how they may be addressed.
 \item As a contribution for researchers acquainted with the VPR problem, it examines the intricacies of different VPR problem types regarding input (database or query set), data processing (online or offline) and output (one or multiple matches per query image). The tutorial also discusses the subtleties behind the evaluation of VPR algorithms, e.g., the evaluation of a VPR system that has to find \textit{all} matching database images per query, as opposed to just a \emph{single} match.
\end{enumerate}
Practical code examples in Python illustrate to prospective practitioners and researchers how VPR is implemented and evaluated. The corresponding source code is available online, along with a list of other open-source implementations from the literature: \url{https://github.com/stschubert/VPR_Tutorial}.
The link also provides a Jupyter notebook written in Python that guides through a basic VPR pipeline. It allows users to experiment with additional image descriptors, benchmark datasets, and evaluation metrics.

The following \autoref{sec:vpr_basic} provides a basic introduction to the VPR problem. \autoref{sec:tutorial_structure} then outlines the structure of the remainder of this tutorial.

\subsection{The basics of Visual Place Recognition (VPR)}\label{sec:vpr_basic}
VPR involves matching one or multiple image sets in order to determine which images show the same places in the world.
These image sets are typically recorded with a mobile device such as a cell phone or an AR/VR headset, or with a camera mounted on a variety of platforms, such as a robot, uncrewed aerial vehicle, car, bus, train, bicycle, or boat.

Essentially, VPR is an image retrieval problem where the context is to recognize previously seen places. This context provides additional information and structure beyond a general image retrieval setup. Many VPR methods exploit the context to match images of the same places in a wide range of environments, including those with significant appearance and viewpoint differences. For example, one piece of additional information that is often exploited is that \emph{consecutive} images taken by a camera mounted on a car will depict spatially close places in the world.

\autoref{fig:basic_pipeline} provides an overview of the typical steps and components of a basic VPR pipeline. 
Given a reference set composed of database images $\mathbf{I}_i\in DB$ of known places, and one or multiple query images $\mathbf{I}_j\in Q$, the goal is to find matches between these two sets, i.e.,~those instances where image~$j$ from the query set shows the same place as image~$i$ from the database.
To find these matches, it is essential to compute one or multiple descriptors $\mathbf{d}$ for each image -- these descriptors should be similar for images showing the same place and dissimilar for different places. A descriptor is typically represented as a numerical vector (e.g.,~128-D or 4,096-D). 
Conceptually, we can think of a matrix $\mathbf{S}$ of all pairwise descriptor similarities $s_{ij}$ between the database and query images as the basis for deciding which images should be matched. 
In practice, we must carefully choose the algorithms used to compute and compare the image descriptors $\mathbf{d}$, taking into account the specific challenges and context of the VPR problem at hand. 
The remainder of this tutorial will provide more detail on these aspects and discuss the VPR problem from a broader theoretical and practical perspective.

\subsection{The VPR problem and its details as reflected in this tutorial}\label{sec:tutorial_structure}
\autoref{sec:history} of this tutorial will outline the relevance and history of the VPR problem, as well as its relation to other areas, particularly its importance for topological Simultaneous Localization And Mapping (SLAM), where the database $DB$ corresponds to the set of previously visited places in the map. In fact, one of the original drivers for VPR research was the generation of loop closures for SLAM systems, that is, 
recognizing a place when revisiting it (e.g., in a loop) and tying the current observation with that already in the map (i.e., closure)~\cite{tsintotas2022revisiting}.
One of the earliest examples of such a topological SLAM system is FAB-MAP~\cite{cummins08}, also referred to as `appearance-only SLAM', where loop closure generation is based on appearance only (i.e., images), thus different from 3D/metric SLAM systems such as ORB-SLAM~\cite{MurArtal15} where the map and the visual landmarks are expressed in 3D.

The definition of a ``\textit{place}'' is an integral aspect of VPR.
In this tutorial, we follow the definition that two images must have some visual overlap, i.e., shared image content like same buildings, to be considered as ``taken at the same place''~\cite{Garg2021}. This definition allows to subsequently estimate the camera transformation between matched images for tasks like visual localization, mapping, or SLAM -- indeed, the required amount of visual overlap depends on the specific application.  %
We note that an alternative definition used in particular by some researchers~\citemore{yin2022general}{Weyand2016} is that two places are matching purely based on their position, without taking the orientation, and in turn visual overlap, into account. 
\autoref{sec:taxonomy} will present different applications for VPR and discuss the various subtypes of VPR problems that arise from variations in the available input data, the required data processing, and the requested output. 

VPR algorithms are often tailored to the particular properties of an application. \autoref{sec:vpr_problem} will provide details on a \textit{generic VPR pipeline} that serves as a common basis for diverse practical settings and their unique characteristics. From this section onward, this tutorial includes practical code examples in Python. %

It is important to note that not all VPR algorithms address the same VPR problem, e.g., regarding the requested number of image matches per query. This is particularly critical when it comes to evaluating and comparing the performance of different VPR algorithms. \autoref{sec:evaluation} explains and discusses the evaluation pipelines that consider various datasets, ground truth subtleties, and different performance metrics. 

The properties of the underlying data have a significant impact on the difficulty of the resulting VPR problem and the suitability of a particular algorithm.
\autoref{sec:challenges} will discuss challenges such as severe appearance changes due to varying illumination or weather conditions, large viewpoint changes between images of the same place, and perceptual aliasing, i.e.,~the challenge that images taken at two distinct places can appear remarkably similar. This section will also present common ways of addressing these challenges to improve robustness, performance, runtime and memory efficiency.
These approaches include methodological extensions of the general purpose pipeline that partially build upon a robotic context (e.g., with image sets recorded as videos along trajectories) where VPR differs from pure image retrieval.
This often allows the exploitation of additional knowledge and information such as spatio-temporal sequences (i.e., consecutive images in the database $DB$ and query $Q$ are also neighboring in the world) or intra-set similarities (i.e., similarities \textit{within} $DB$ or $Q$).

%% file: 2-History.tex
\section{History, Relevance and Related Areas}
\label{sec:history}
    Visual Place Recognition (VPR) research can be traced back to advances in visual SLAM, visual geo-localization, and image retrieval applied to images of places~\cite{Durrant-Whyte2006}. %
    In the robotics literature, VPR has historically been called loop closure detection and was mainly used for this purpose for visual SLAM~\cite{Durrant-Whyte2006}. VPR gained more prominence in the field as the
    earlier metric SLAM methods based on global and local bundle adjustment techniques could only handle limited-size environments, thus paving way for topological SLAM techniques based on bag-of-words approaches\ifarxiv~\cite{galvez2012bags}\fi, such as FAB-MAP~\cite{cummins08}. In addition to its relevance within SLAM pipelines, VPR also remains a crucial component of localization-only pipelines where the map is available a priori.%
    
    Early VPR research primarily focused on place recognition under constant or slightly varying environmental conditions. Addressing appearance changes due to more severe condition changes, such as day-night cycles or seasonal shifts, emerged in the late 2000s. These methods relied for example on local feature matching~\cite{Valgren2010} or on continuously updating appearance-based maps~\cite{dayoub2008adaptive}. Since then, research on VPR under challenging conditions has steadily increased, for example tackling the challenging day-night shift~\cite{seqSLAM}.
    Recent works make heavy use of datasets with condition changes that have appeared since 2012~\cite{Schubert2021,Masone2021}.
    In 2014, the use of deep learning for VPR~\cite{zetao2014} emerged as a way to handle challenging data and has since proven effective in changing environments~\cite{sunderhauf2015performance}.
    In addition to images and image descriptors, VPR research has also explored the use of additional information, such as sequences, intra-set similarities, weak GPS signals, or odometry, to improve performance~\cite{Schubert2021b}.

    In terms of the relationship between VPR and other fields, we recommend the following tutorials: \cite{Durrant-Whyte2006} provides an overview of probabilistic SLAM and includes a section on loop closure detection, although a lot of progress has been made in VPR as this tutorial was published more than 15 years ago. \cite{tsintotas2022revisiting} specifically investigates the loop closure problem in SLAM. \ifarxiv\cite{Aslan2021} provides a practical introduction to SLAM with example code for the Robot Operating System (ROS). \fi \cite{scaramuzza2011visual} discusses visual odometry, which involves estimating the ego-motion of an agent based on visual input. Visual odometry is thus complementary to VPR, and can be combined with VPR to detect loop closures when building a SLAM system.
    It is important to note, however, that the scope of this tutorial is limited to providing an accessible introduction to VPR and its core concepts. Aspects such as the integration of VPR methods into a complete SLAM or re-localization system are beyond the scope of this tutorial and would require discussing many additional aspects, such as batch optimization, which are not directly related to VPR.
Beyond loop-closure detection, VPR is necessary if global position sensors such as global navigation satellite systems (GNSS) like GPS, Galileo, or BeiDou are not available or are inaccurate.
In urban environments, buildings or other structures can lead to ``urban canyons'' that block line-of-sight satellite signals, causing occlusions that prevent a GNSS receiver from obtaining accurate position information. In addition to occlusions, reflections of GNSS signals off buildings and other structures, so called non-line-of-sight signals, can further hinder the GNSS accuracy. This issue is not limited to urban environments, as similar occlusions and reflections can occur in natural environments, such as in valleys or canyons. Similarly, indoor environments and caves also hinder GNSS due to the absorption or reflection of satellite signals by walls.

Alternatively, VPR can serve as a redundant component in autonomous systems for fault tolerance and general GNSS outages, such as satellite service disruptions, degradation, or position/time anomalies.
It is worth noting that all GNSS systems can potentially be hacked or blocked for non-military use by a central authority.
Other systems may not be equipped with a GNSS receiver due to cost or security concerns.
In the case of robotic extraterrestrial missions, installing a GNSS system may be too expensive or time-consuming.

%% file: 3-CategoriesUsecases.tex
\section{VPR problem categories and use cases}
\label{sec:taxonomy} 
In the localization and mapping literature, VPR has been used in different ways depending on three key attributes of its formulation: the \textit{input}, which deals with how the reference and query images are made available (i.e., single-session vs.~multi-session); \textit{data processing}, that defines the mode of operation (i.e., online vs.~batch); and \textit{output}, that determines the kind of expected output (i.e., single-best-match vs.~multi-match).
The following \autoref{sec:vpr_categories} explains these problem categories in more detail. 
\autoref{sec:vpr_use_cases} then presents different VPR use cases using these categories.
\autoref{tab:criteria} summarizes these use cases, along with their required input and data processing.
Note that there might be exceptions and deviations from these categories, such as \cite{Vysotska2019} that uses multiple disjoint sequences as reference.
However, we believe that the proposed taxonomy serves as a good starting point for future research to organize the various VPR use cases.

\subsection{VPR problem categories}\label{sec:vpr_categories}
We distinguish three main dimensions along which VPR problems can vary, creating different VPR problem categories or subtypes:
\begin{enumerate}[leftmargin=0.46cm]
    \item \textbf{Input -- single-session VPR vs.~multi-session VPR}:
    \textit{Are there two separate input sets, one for the database $DB$ and one for the query $Q$, or is it a single set that is compared to itself?}
    Single-session VPR is the matching of images within a single set of images, so that the query set $Q$ equals the database $DB$ (i.e., $Q=DB$). A practical consideration in this case is the suppression of matches with recently acquired images -- while full SLAM systems typically rely on a motion model for such suppression, standalone VPR systems often use heuristics.
    In contrast, multi-session VPR is the matching of the two disjoint image sets (i.e.,~$DB\cap Q=\varnothing$), which were recorded at different times (e.g., summer and winter) or by different platforms (e.g., mobile robot and cell phone).

    \item \textbf{Data processing -- online VPR vs.~batch VPR}: 
    \textit{Are the images available and processed individually, one after the other, or are they all available in a single batch from the beginning?} 
    Online VPR has to deal with a growing set $Q$ (i.e., $Q\neq\const$) and a set $DB$ that is either given (i.e., $DB=\const$) or also growing (i.e., $DB\neq\const$).
    In contrast, batch VPR can build upon the full sets $Q$ (i.e., $Q=\const$) and $DB$ (i.e., $DB=\const$).
    Growing image sets in the case of online VPR limit the number of viable methods.
    For example, approaches like descriptor standardization~\cite{Schubert2020} based on the statistics of \textit{all} image descriptors or similarity matrix decomposition~\cite{ho2007detecting} cannot be used without modifications. This is further discussed in \autoref{sec:challenges}.
    Note that for single-session VPR, the difference between a growing or constant query set pertains to whether VPR is being performed online or in batch mode.
    
    \item \textbf{Output -- single-best-match VPR vs.~multi-match VPR}: 
    \textit{Is the intended output for a query image a single image from the database that shows the same place, or do we request all images of this place?}
    Single-best-match VPR only returns the best matching database image $I^*_i\in DB$ per query image $I_j\in Q$.
    In contrast, the aim in multi-match VPR is finding \textit{all} matching database images for each query image. In practice, the difference between single-best-match VPR and multi-match VPR often boils down to finding either the maximum similarity between a query and all database images or all similarities above a certain threshold, as shown in~\autoref{sec:output}. Identifying all matching images is often more challenging than finding only one correct match, as it requires an explicit decision for each database image whether it shows the same place as the query image or not~\cite{Schubert2021}.
\end{enumerate}

\noindent
Let us illustrate these problem categories with the example of determining the rough pose [x, y, heading, floor] of a cell phone in a building, e.g., to guide persons to desired places.
To achieve this, we first need to map the building before the person can use their cell phone to localize in the building.
For this first step of mapping the building, we could use a manually controlled mobile robot equipped with a camera to collect a query set~$Q^\text{mapping}$ of images together with some additional sensor data like odometry.
Given these images of all places, we can run a mapping algorithm that processes all images and other data to obtain a metric map of the building, which associates all images in $Q^\text{mapping}$ with metric poses.
Part of this mapping is a \textit{single-session batch multi-match VPR} for loop closure detection that compares the whole image set $Q^\text{mapping}$ (batch VPR) to itself (single-session VPR) to find \textit{all} loop closures for each image (multi-match VPR).
Here, batch processing the whole set $Q^\text{mapping}$ allows the application of computationally expensive but accurate algorithms.

\begingroup
\renewcommand*{\arraystretch}{1.2}
\begin{table}[t]
    \setlength{\tabcolsep}{2.1pt}
    \centering
    \normalsize
    \caption{Combinations of the VPR \textit{input} and \textit{data processing} categories (cf.~\autoref{sec:vpr_categories}) with corresponding use cases (cf.~\autoref{sec:vpr_use_cases}). In single-session VPR, there is a single input set that is compared to itself. This is for example the case in online SLAM, where this set grows while the robot is exploring its environment (online VPR), and in mapping, where pre-recorded data is processed in a batch manner. In the case that the input consists of two sets, the database $DB$ and the query set $Q$ (multi-session VPR), one can again distinguish the case that data needs to be processed online as it is being collected (online VPR) or in a batch as in multi-session mapping. In multi-session online VPR, $DB$ can be either growing (as in multi-robot mapping) or fixed (as in visual (re-)localization).}
    \label{tab:criteria}
        \begin{tabular}{cP{1.7cm}||c|c|c}
            && \multicolumn{3}{c}{\textbf{Data processing}} \\
            && \multicolumn{2}{c|}{\textbf{Online VPR}} & \textbf{Batch VPR} \\ \hline\hline
            \parbox[t]{3mm}{\multirow{5}{*}{\rotatebox[origin=c]{90}{\textbf{Input}}}}&\textbf{Single-session VPR} & \multicolumn{2}{c|}{\multirow{3}{*}{Online SLAM}} & \multirow{3}{*}{Mapping} \\ \cline{2-5}
            &\multirow{3}{1.7cm}{\centering \textbf{Multi-session VPR}} & \boldmath\textbf{$DB$ grows} & \boldmath\textbf{$DB$ const.} & \multirow{3}{*}{\shortstack{Multi-Session\\[0.1cm]Mapping}} \\ \cline{3-4}
            && Multi-Robot & Visual (Re-) &  \\
            && Mapping & Localization &  %
        \end{tabular}%
\end{table}
\endgroup

\begin{figure}[t]
    \centering
    \includegraphics[width=0.9\linewidth]{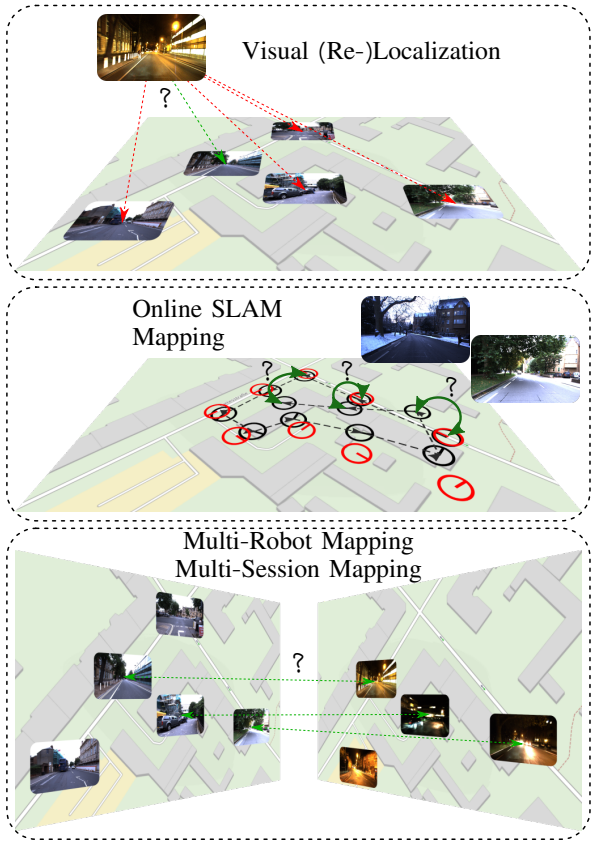}
    \vspace{-0.1cm}
    \caption{Overview of Visual Place Recognition use cases. Top: VPR is used to select a small set of candidate images from the database for visual (re-)localization, which are processed with computationally expensive 6-DoF pose estimation methods. Middle: VPR can be used to detect loop closures in a SLAM pipeline (green arrows) or to re-localize after mislocalization or if the robot was moved (kidnapped robot). Bottom: In multi-session mapping, VPR can again be used for loop closure detection, but this time in a batch manner where all images are known in advance. In multi-robot mapping, VPR is used to merge maps by detecting places that have been visited by multiple robots.
    (Map data from \href{https://openstreetmap.org/copyright}{OpenStreetMap})
    }
    \label{fig:use_cases}
    \vspace{-0.2cm}
\end{figure}

After mapping (potentially years later), the second step is the actual localization of a cell phone using its camera stream.
When localizing, we treat the robot's mapping query set~$Q^\text{mapping}$ as database~$DB^\text{loc}$ and compare it to query images~$Q^\text{loc}$ from a cell phone's camera.
To determine the location of a cell phone, a \textit{multi-session online single-best-match VPR} can be used that compares the stream of query images $Q^\text{loc}$ to the fixed database~$DB^\text{loc}$ (multi-session VPR) online (online VPR) to find the best matching database image (single-best-match VPR) with its corresponding pose information.

In summary, VPR can be used for a variety of different use cases, as discussed in more detail in the following \autoref{sec:vpr_use_cases} and shown in \autoref{tab:criteria}.
Here, each use case typically requires a certain combination of the \textit{input} (single-session/multi-session VPR) and \textit{data processing} (online/batch VPR) VPR categories.
The choice of the \textit{output} category (single-best-match/multi-match VPR) also depends on the use case, and in particular on the algorithm that is used after VPR.
For example, for pure (re-)localization~\cite{Sattler2018}, one may only need a single best match, so that the choice of the output category depends mainly on the use case, while for graph-based SLAM, the required output category also depends on the post-processing after VPR, as explained in the following.
In graph-based SLAM~\cite{Durrant-Whyte2006}, each node encodes the pose of an image in $Q$.
The corresponding edges of connected nodes represent the transformation between them.
An edge can be established either between temporally consecutive nodes (using the odometry) or between nodes that were identified as loop closure by VPR.
Here, single-best-match VPR could be used to match and fuse two nodes which correspond to the same place to represent each place always by only one node.
Alternatively, multi-match VPR could be used to create multiple edges between all existing nodes of the same place. This is particularly helpful if we cannot guarantee that there is a single node for each place in the graph, or if we perform a batch optimization of the poses using a robust optimization approach that can benefit from the additional information provided by multiple matches while handling potential outlier matchings.

\subsection{VPR use cases}\label{sec:vpr_use_cases}
VPR is a key component in a variety of robotic applications, including autonomous driving, agricultural robotics, and robotic parcel delivery, as well as in the creation of a metaverse. Some common tasks that VPR is used for include:
\begin{enumerate}[leftmargin=0.46cm]

    \item \textbf{Candidate selection for 6~DoF visual localization}~\cite{Sattler2018}: 6~Degree Of Freedom (DoF) visual localization (also termed city-scale/natural geo-localization) involves estimating the 6D pose (position and orientation) of a camera in a particular environment. \textit{Multi-session online VPR with fixed $DB$} is used to select candidates $\mathbf{I}_i\in DB$ that have the highest similarity to the current query images $\mathbf{I}_j\in Q$ (cf.~\autoref{fig:use_cases}, top).
    These candidates can then be used for a computationally intensive 6D pose estimation using local image descriptors and more complex algorithms, which would be infeasible for the complete $DB$ set.
    \ifarxiv For example, in \cite{Dusmanu2019} the place recognition method NetVLAD~\cite{Arandjelovic_2016_CVPR} was used for candidate selection before performing pose estimation with local descriptors.\fi
    
    \item \textbf{Loop closure detection and re-localization for online SLAM}~\cite{Durrant-Whyte2006}: Online SLAM is used to estimate the current pose of a camera while creating a map of the environment at the same time.
    \textit{Single-session online VPR} is used for loop closure detection (i.e., the recognition of previously visited places), as shown in \autoref{fig:use_cases} (mid) to compensate for accumulated errors in odometry data and create a globally consistent map. It is also used for re-localization in the event of mislocalization or if the camera/robot was moved by hand (known as the kidnapped robot problem).
    
    \item \textbf{Loop closure detection for mapping}~\citemore{Durrant-Whyte2006}{Frese2010}: Mapping (also full SLAM or offline SLAM) involves estimating the entire path at once to generate a map. 
    This allows for the use of \textit{single-session batch VPR} for loop closure detection (cf.~\autoref{fig:use_cases}, mid), which is based on slower but more robust algorithms that run on powerful hardware.
    
    \item \textbf{Loop closure detection for multi-session mapping}~\cite{McDonald2011}: Multi-session mapping combines the results of
    multiple SLAM missions performed repeatedly over time in the same environment.
    \textit{Multi-session batch VPR} is used to find shared places between the individual maps of all missions for map merging (cf.~\autoref{fig:use_cases}, bottom). Alternatively, \textit{multi-session online VPR with a given DB} can be used to detect previously mapped areas (potentially for loop closure) and include unseen areas of the map in real-time.
    
    \item \textbf{Detection of shared places for multi-robot mapping}~\cite{Cieslewski2018}: Multi-robot mapping (also termed decentralized SLAM) involves the distributed mapping of an environment using multiple robots.
    Here, \textit{multi-session online VPR with a growing $DB$} is used to find shared places between the individual maps of each robot for subsequent map merging, as shown in \autoref{fig:use_cases} (bottom).
    
\end{enumerate}

\noindent In summary, this section provided an overview of the different problem categories and corresponding subtypes of VPR and discussed common use cases where VPR is applied.

%% file: 4-Pipeline.tex
\section{A generic pipeline for VPR}
\label{sec:vpr_problem}
This section outlines a generic pipeline for Visual Place Recognition (VPR). The steps involved in this pipeline are shown in \autoref{fig:basic_pipeline}.
The inputs to the pipeline are two sets of images $DB$ and $Q$ (these may be the same for single-session VPR, as explained in \autoref{sec:taxonomy}).
The pipeline produces matching decisions, meaning that for each query image $\mathbf{I}_j\in Q$, one or more database images $\mathbf{I}_i\in DB$ can be associated.
The pipeline includes these intermediate steps and components: 1)~computing image-wise descriptors, 2)~pairwise comparing of descriptors to 3)~create a descriptor similarity matrix $\mathbf{S}$, and 4)~making matching decisions.
In the following subsections, we will discuss each of these elements in more detail. Extensions to this generic pipeline that can be used to improve performance and robustness against various challenges are presented in \autoref{sec:challenges}.

\subsection{Inputs: The database and query image sets}\label{sec:input}
To recap, two sets of images serve as the input in a VPR pipeline: the database set $DB$ and a set of current images in the query set $Q$. The $DB$ set, which is also called the reference set, represents a map of known places, and is often recorded under ideal conditions (e.g., sunny), or by a different platform than $Q$ (e.g., a second robot). 
The query set $Q$, on the other hand, is the ``live view'' recorded by a different platform than $DB$ or after $DB$ -- potentially days, months, or even years later. Both sets will have a geographical overlap and share some or all seen places.

There are different VPR problem categories: using just a query set $Q$ (single-session VPR) or using both the $DB$ and $Q$ sets (multi-session VPR).
Also, the image sets can either be specified before processing (batch VPR) or grow during an online run (online VPR).

\autoref{code:dataset_loading} provides example code for loading a dataset with both image sets $Q$ and $DB$, as well as the ground truth matrices $\mathbf{GT}$ and $\mathbf{GT}^{soft}$. Briefly, $\mathbf{GT}$ is a logical matrix that indicates whether corresponding images show the same or different places, while $\mathbf{GT}^{soft}$ is a dilated version of $\mathbf{GT}$ that accounts for image pairs with small visual overlap, avoiding penalization for matches in such cases. We detail these matrices in~\autoref{sec:GT}.

\begin{program}[t]
\begin{mdframed}[style=example]
\begin{minted}[fontsize=\footnotesize,escapeinside=||,mathescape=true]{python3}
# load dataset GardensPoint Walking $\cite{ds_gardenspoint_riverside}$ with two 
# image sets DB and Q and ground truth
from datasets.load_dataset \
    import GardensPointDataset
dataset = GardensPointDataset()
|$DB$|, |$Q$|, |$\mathbf{GT}$|, |$\mathbf{GT}^{soft}$| = dataset.load()
\end{minted}

\par\noindent\rule{\textwidth}{0.4pt}

\noindent Output:
    
\footnotesize
\noindent
$\mathbf{I}_1{\in} DB$: \includegraphics[width=0.35\linewidth]{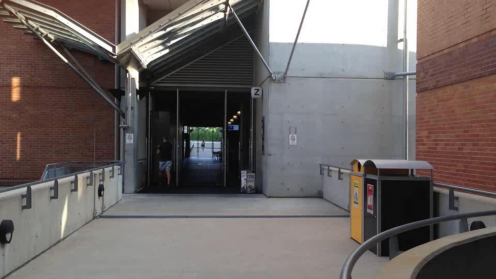}
\hfill
$\mathbf{I}_1{\in} Q$: \includegraphics[width=0.35\linewidth]{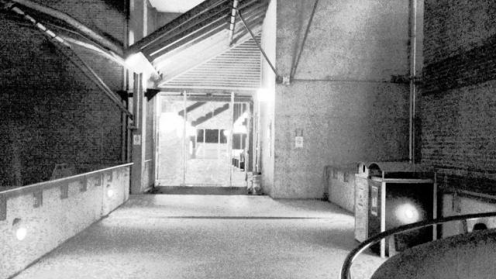}

\vspace{0.3cm}

\noindent
$\mathbf{GT}$: \includegraphics[width=0.38\linewidth]{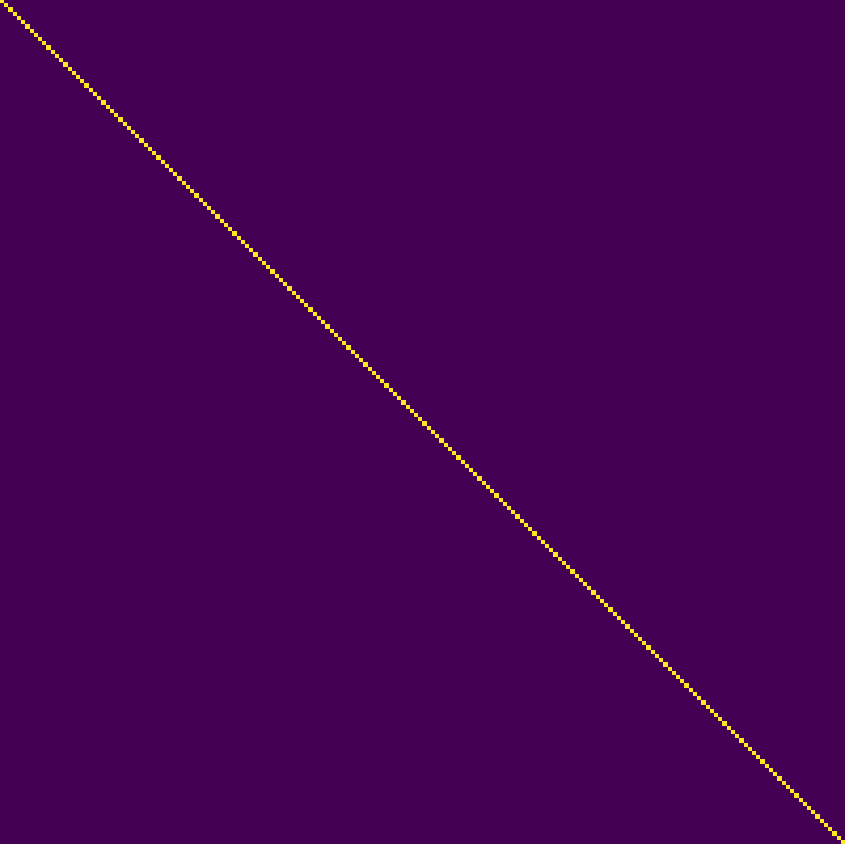}
\hfill
$\mathbf{GT}^{soft}$: \includegraphics[width=0.38\linewidth]{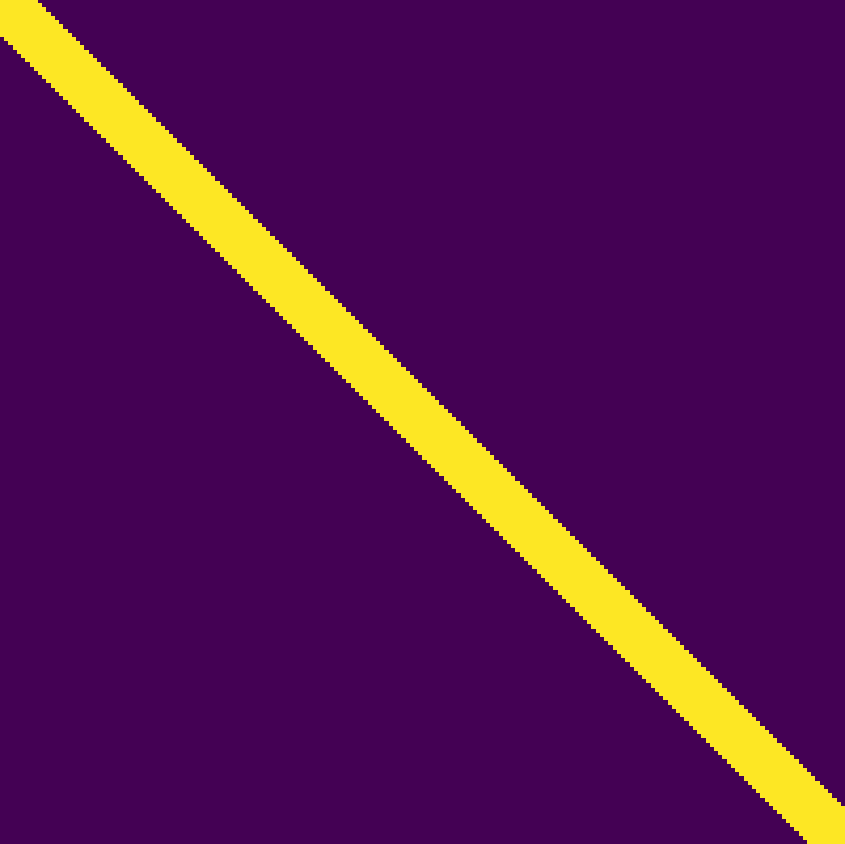}
\end{mdframed}
\caption{In this example, the input to a VPR algorithm are two disjoint image sets: the database~$DB$ and query~$Q$. We load the GardensPoint Walking dataset~\cite{ds_gardenspoint_riverside} and ground-truth information about correspondences. This ground truth only serves for later evaluation and will neither be available nor required when deploying the algorithm.}
\label{code:dataset_loading}
\end{program}

\subsection{Image-wise descriptor computation}\label{sec:desc_comp}
This section describes the process of computing image descriptors, which are abstractions of images that extract features from raw pixels in order to be more robust against changes in appearance and viewpoint (step 1 in \autoref{fig:basic_pipeline}, see also \autoref{code:descriptor_computation}).
The tutorial covers two primary types of image descriptors:
\begin{enumerate}[leftmargin=0.46cm]
    \item \textbf{Holistic descriptors} (also called global descriptors) represent an image $\mathbf{I}_i\in DB, Q$ with a single vector $\mathbf{d}_i\in\mathbb{R}^{d}$ (cf.~\autoref{code:descriptor_computation}). This allows for efficient pairwise descriptor comparisons with low runtimes. 
    Note that when exhaustive k-nearest neighbor search (kNN) is used to obtain the nearest neighbors for a candidate selection of similar database descriptors, the execution time scales linearly with both the descriptor dimension and the number of images contained in the database.

    \item \textbf{Local descriptors} encode an image $\mathbf{I}_i$ with a set $D_i=\{\mathbf{d}_k\mid k=1,\dots, K\}$ of vectors $\mathbf{d}_k\in \mathbb{R}^d$ at $K$ regions of interest. They often provide better performance than holistic descriptors, but require computationally expensive methods for local feature matching like a left-right check (also termed mutual matching)\ifarxiv~\cite{Neubert2021}\fi, a homography estimation\ifarxiv~\cite{delf}\fi, a computation of the epipolar constraint\ifarxiv~\cite{Valgren2010}\fi, or deep-learning matching techniques\ifarxiv, e.g., SuperGlue \cite{superglue}\fi.
    Therefore, local descriptors are typically used in a hierarchical pipeline, where first the %
    holistic descriptors are used to retrieve the top-$K$ matches, which are then re-ranked using local descriptor matching.
\end{enumerate}
The abstraction of the VPR pipeline in terms of holistic and local descriptors serves as the foundation for many localization, mapping, and SLAM solutions. Alternative approaches include place classification~\cite{Weyand2016}, regional descriptors~\cite{Zhang2021}, and incremental bags of binary words~\cite{garcia2018ibow}. Furthermore, in \autoref{sec:challenges} we list common shortcomings of this VPR pipeline and ways to address them.

To convert a set of local descriptors from a single image into a holistic descriptor, one can use \textbf{local feature aggregation} methods like Bag of Visual Words (BoVW)\ifarxiv~\cite{Sivic2003}\fi, Vector of Locally Aggregated Descriptors (VLAD)\ifarxiv~\cite{Jegou2010}~\else~\fi or Hyper-Dimensional Computing (HDC) \cite{Neubert2021}. In a hierarchical pipeline, this allows a local descriptor to be used for both candidate selection (after aggregation) and verification (with the raw local descriptors).

As the descriptor computation is one of the first steps in a pipeline for VPR, it has a significant impact on the performance of subsequent steps and the overall performance of the VPR system.
The algorithm used to obtain the descriptors determine how well the descriptors are suited for a specific environment, the degree of viewpoint change, or the type of environmental condition change.
For example, CNN-based holistic descriptors like AlexNet-conv3 \cite{sunderhauf2015performance} \ifarxiv or HybridNet \cite{hybridnet} \fi perform well in situations with low or negligible viewpoint changes, but perform poorly with large viewpoint changes.
On the other hand, VLAD-based \citemore{}{Jegou2010} descriptors like NetVLAD \cite{Arandjelovic_2016_CVPR} tend to perform better in settings with large viewpoint changes.

Additionally, the specific training data of deep-learned descriptors affect the performance in different environments.
For example, some descriptors may perform better in urban environments, while others may be more effective in natural environments \citemore{}{Garforth2020} or in specific geographic regions such as Western cities~\cite{warburg2020mapillary}.

\begin{figure*}[t]
    \centering
    \includegraphics[width=1\linewidth]{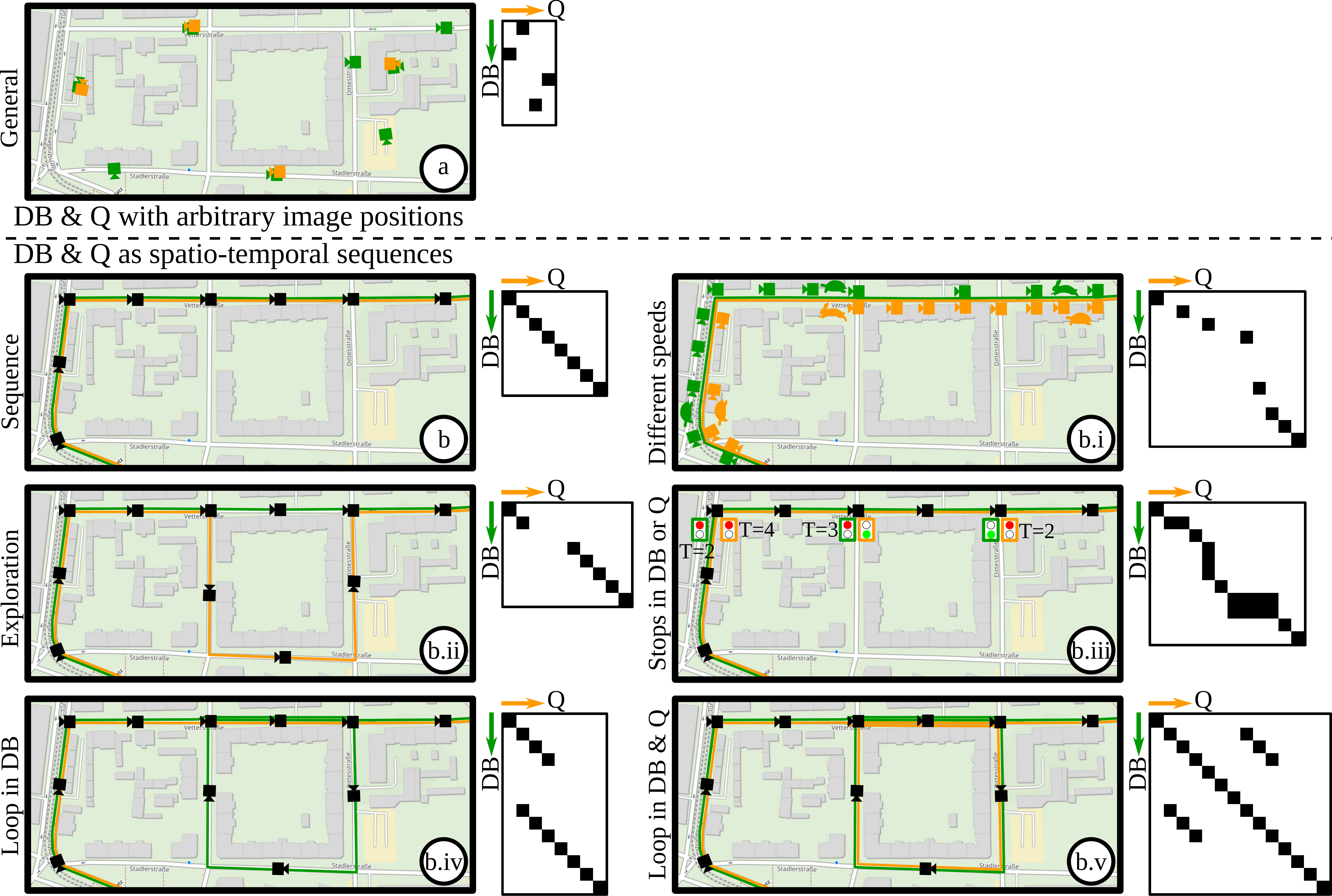}
    
    \caption{Relation between the similarity matrix $\mathbf{S}$ and the trajectory during the database and query run. Green and orange cameras depict images in $DB$ and $Q$, respectively. Green and orange lines indicate that images were recorded as video along a trajectory (also called spatio-temporal sequence). In b.i), a rabbit/turtle indicate fast/slow speeds when traversing the route, and similarly in b.iii) traffic lights indicate stops in $Q$ (T=2), $DB$ (T=3) or both (T=\{2,4\}) for T time steps.}
    \label{fig:relation_traj_S}
\end{figure*}

\begin{program}[t]
\begin{mdframed}[style=example]
\begin{minted}[fontsize=\footnotesize,escapeinside=||,mathescape=true]{python3}
# compute holistic HDC-DELF descriptors $\cite{Neubert2021}$
from feature_extraction. \
    |\textbf{\textcolor{blue}{feature\dunderline{0.39pt}{\hspace{0.171cm}}extractor\dunderline{0.39pt}{\hspace{0.171cm}}holistic}}| import HDCDELF
feature_extractor = HDCDELF()
|$\mathbf{D}^{DB}$| = feature_extractor.compute_features(DB)
|$\mathbf{D}^Q$| = feature_extractor.compute_features(Q)
\end{minted}
\end{mdframed}
\caption{The main source of information about image correspondences are image descriptors. Since holistic image descriptors allow for efficient pairwise descriptor comparisons, we compute a holistic HDC-DELF~\cite{Neubert2021} descriptor for each image (step~1 in \autoref{fig:basic_pipeline}).}
\label{code:descriptor_computation}
\end{program}

\subsection{Descriptor similarity between two images}
To compare the image descriptors of two images, a measure of similarity or distance must be calculated (see step 2 of \autoref{fig:basic_pipeline} and \autoref{code:descriptor_comparison}). This process compares the descriptors $\mathbf{d}_i$ and $\mathbf{d}_j$ (holistic) or $D_i$ and $D_j$ (local) of images $i$ and $j$.
Note that similarity $s_{ij}$ and distance $dist_{ij}$ can be related through inversely proportional functions such as
\begin{align}
    s_{ij}&=-dist_{ij}\,,\\
\intertext{or the reciprocal}
    s_{ij}&=\frac{1}{dist_{ij}} \ .
\end{align}

Holistic descriptors can be compared more efficiently than local descriptors, as they only require simple and computationally efficient metrics like the cosine similarity
\begin{equation}
    s_{ij} = \frac{\mathbf{d}_i^T\cdot \mathbf{d}_j}{\|\mathbf{d}_i\|\cdot\|\mathbf{d}_j\|},
\end{equation}
or the negative Euclidean distance
\begin{equation}
    s_{ij} = -\|\mathbf{d}_i-\mathbf{d}_j\|.
\end{equation}
In contrast, comparing local descriptors requires more complex and computationally expensive algorithmic approaches, as previously mentioned in \autoref{sec:desc_comp}. 

\begin{program}[t]
\begin{mdframed}[style=example]
\begin{minted}[fontsize=\footnotesize,escapeinside=||,mathescape=true]{python3}
# compare all descriptors using cosine similarity
# normalize descriptors
|$\mathbf{D}^{DB}$| = |$\mathbf{D}^{DB}$| / 
      np.linalg.norm(|$\mathbf{D}^{DB}$|, axis=1, keepdims=True)
              
|$\mathbf{D}^Q$| = |$\mathbf{D}^Q$| /
      np.linalg.norm(|$\mathbf{D}^Q$|, axis=1, keepdims=True)
# dot product between descriptors using matrix 
# multiplication
|$\mathbf{S}$| = np.matmul(|$\mathbf{D}^{DB}$|, |$\mathbf{D}^Q$|.transpose())
\end{minted}
\par\noindent\rule{\textwidth}{0.4pt}

\noindent Output:

\centering
\noindent
$\mathbf{S}$: \includegraphics[width=0.6\linewidth]{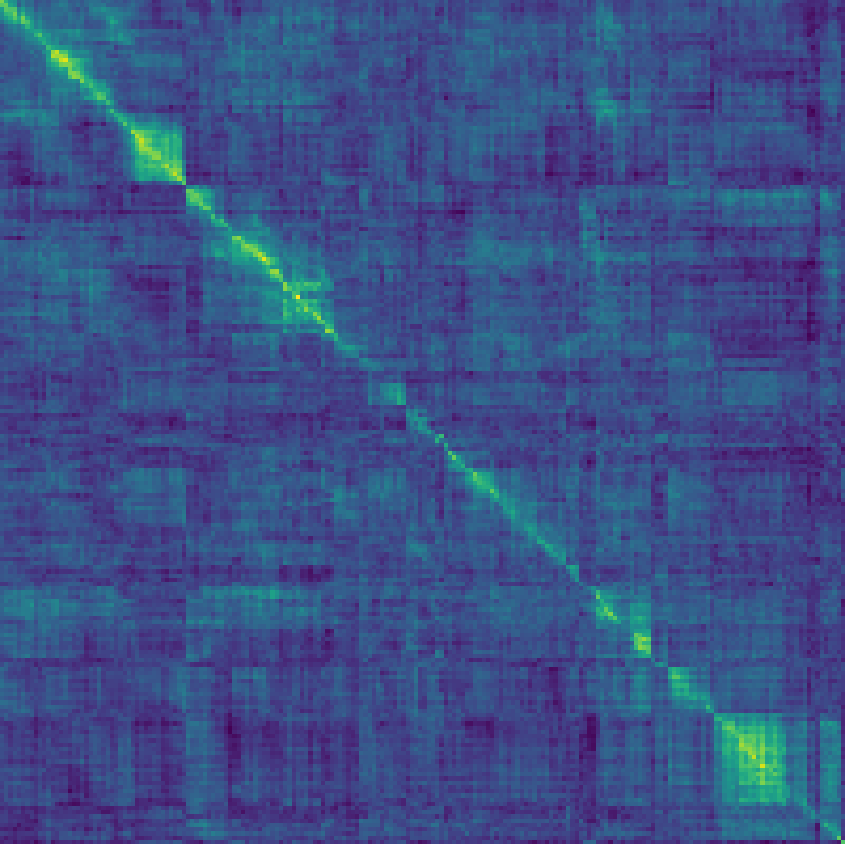}
\end{mdframed}
\caption{To compare database and query descriptors to obtain the descriptor similarities $\mathbf{S}$ (steps~2 and~3 in \autoref{fig:basic_pipeline}), we use their cosine similarity (computed by the inner product of the normalized descriptor vectors). Although we might not want to compute the full similarity matrix $\mathbf{S}$ of all possible pairs in a large-scale practical application, it can be useful for visual inspection purposes.}
\label{code:descriptor_comparison}
\end{program}

\subsection{The pairwise similarity matrix \texorpdfstring{$\mathbf{S}$}{S}}\label{sec:S}
The pairwise descriptor similarity matrix $\mathbf{S}$ is a key component of VPR. As shown in step 3 of \autoref{fig:basic_pipeline}, $\mathbf{S}$ contains all calculated similarities $s_{ij}$ between the descriptors of images in the database and query sets. In single-session VPR, $\mathbf{S}$ has dimensions $|Q|\times|Q|$, while in multi-session VPR, $\mathbf{S}$ has dimensions ${|DB|\times|Q|}$.
Depending on the approach used, $\mathbf{S}$ may be dense (if all descriptors are compared) or sparse (if only a subset of descriptors is compared using approximate nearest neighbor search \citemore{}{Li2020} or sequence-based comparison strategies\ifarxiv~\cite{Schubert2021a}\fi).

The overall appearance of $\mathbf{S}$ is influenced by the camera's trajectories during acquisition of $Q$ and $DB$, as illustrated in \autoref{fig:relation_traj_S}.
The pattern of high similarities within $\mathbf{S}$ can have a significant impact on the performance of the VPR pipeline, and may enable or hinder the use of certain algorithmic steps for performance improvements.
The following relations between camera trajectories and the appearance of $\mathbf{S}$ can be observed (cf.~\autoref{fig:relation_traj_S} for corresponding examples in a map):
\begin{enumerate}[labelindent=0pt,labelwidth=\widthof{a)},itemindent=0em,leftmargin=!]
\item[a)] \textbf{General}: If images in $DB$ and $Q$ are recorded at arbitrary positions without a specific order, there are no discernible patterns in $\mathbf{S}$. This is typical for general visual localization and global geo-localization.

    \item[b)] \textbf{Sequence}: If images in $DB$ and $Q$ are recorded along trajectories as spatio-temporal sequences (i.e., consecutive images are also neighbors in the world), continuous lines of high similarities may be observed in $\mathbf{S}$.
    This setup is typical for many robotic tasks, including online SLAM, mapping, and multi-robot/multi-session mapping (\autoref{sec:taxonomy}). In this setup, sequence-based methods can be used for performance improvements (cf.~\autoref{sec:challenges}).
    The camera's trajectories can affect $\mathbf{S}$ in the following ways:

\begin{enumerate}[labelindent=0pt,labelwidth=\widthof{b.)},itemindent=0em,leftmargin=!]
    \item[i)] \textbf{Speed}: If the camera moves at the same speed in the same locations in $DB$ and $Q$, lines of high similarities with $45^\circ$ slope will be observed. Otherwise, the slope will vary. %
    \item[ii)] \textbf{Exploration}: If a place shown in a query image $Q$ is not present in $DB$, the line of high similarities will be discontinuous.
    \item[iii)] \textbf{Stops}: If the camera stops temporarily (zero velocity) during either the database run or the query run, it will result in multiple consecutive matches in the other set.
    \begin{itemize}
        \item {\boldmath\textbf{\boldmath Stops in $DB$}}: Stops in the database run will result in a vertical line (within the same column) of high similarities in $\mathbf{S}$.
        \item {\boldmath\textbf{\boldmath Stops in $Q$}}: Stops in the query run will result in a horizontal line (within the same row) of high similarities in $\mathbf{S}$.
        \item {\boldmath\textbf{\boldmath Stops in $DB$ \& $Q$}}: If the camera stops in both the database run and query run at the same place, a block of high similarities will be observed in $\mathbf{S}$.
    \end{itemize}
    \item[iv)] \textbf{Loops in DB}: Loops in $DB$ can result in multiple matching database images for a single query image in $Q$.
    Unlike stops, the multiple matching images due to a loop are not consecutive in their image set.
    \item[v)] \textbf{Loops in DB \& Q}: Loops in $DB$ and $Q$ can result in additional matching query images for a single database image in $DB$. This results in a more complex structure of high similarities in $\mathbf{S}$.
\end{enumerate}
\end{enumerate}

\subsection{Output: Matching decisions}\label{sec:output}
The output of a VPR system is a set of matching decisions $m_{ij}\in \mathbf{M}$ (step 4 in \autoref{fig:basic_pipeline} and \autoref{code:finding_matches}) with $\mathbf{M}\in\mathbb{B}^{|Q|\times|Q|}$ (single-session VPR) or $\mathbf{M}\in\mathbb{B}^{|DB|\times|Q|}$ (multi-session VPR) that indicate whether the $i$-th database/query image and the \mbox{$j$-th} query image show the same place ($m_{ij}=true$) or different places ($m_{ij}=false$).
Existing techniques for matching range from choosing the best match per query or a simple thresholding of the pairwise descriptor similarities $s_{ij}\in \mathbf{S}$ to a geometric verification with a comparison of the spatial (using e.g.,~the epipolar constraint) or semantic constellation of the scene. 
For example in \autoref{code:finding_matches}, $M_1$ is computed by selecting the best matching database image per query image, i.e., the maximum similarity $s_{ij}$ per column in $\mathbf{S}\in\mathbb{R}^{|DB|\times|Q|}$ (single-best-match VPR).
Another example is the computation of $M_2$ in \autoref{code:finding_matches}, where a similarity threshold $\theta$ is applied to $\mathbf{S}$: If $s_{ij}\geq\theta$, the $i$-th and $j$-th images are assumed to show the same place (multi-match VPR).
The next section is concerned with the performance evaluation of these outputs.

\begin{program}[t]
\begin{mdframed}[style=example]
\begin{minted}[fontsize=\footnotesize,escapeinside=||,mathescape=true]{python3}
# match images based on S
from matching import matching

# best matching per query in S for
# single-best-match VPR
|$\mathbf{M}_1$| = matching.best_match_per_query(|$\mathbf{S}$|)

# find matches with S>=thresh using an auto-tuned
# threshold for multi-match VPR
|$\mathbf{M}_2$| = matching.thresholding(|$\mathbf{S}$|, thresh='auto')
\end{minted}
\par\noindent\rule{\textwidth}{0.4pt}

\noindent Output:

\centering
\noindent
$\mathbf{M}_1$: \includegraphics[width=0.4\linewidth]{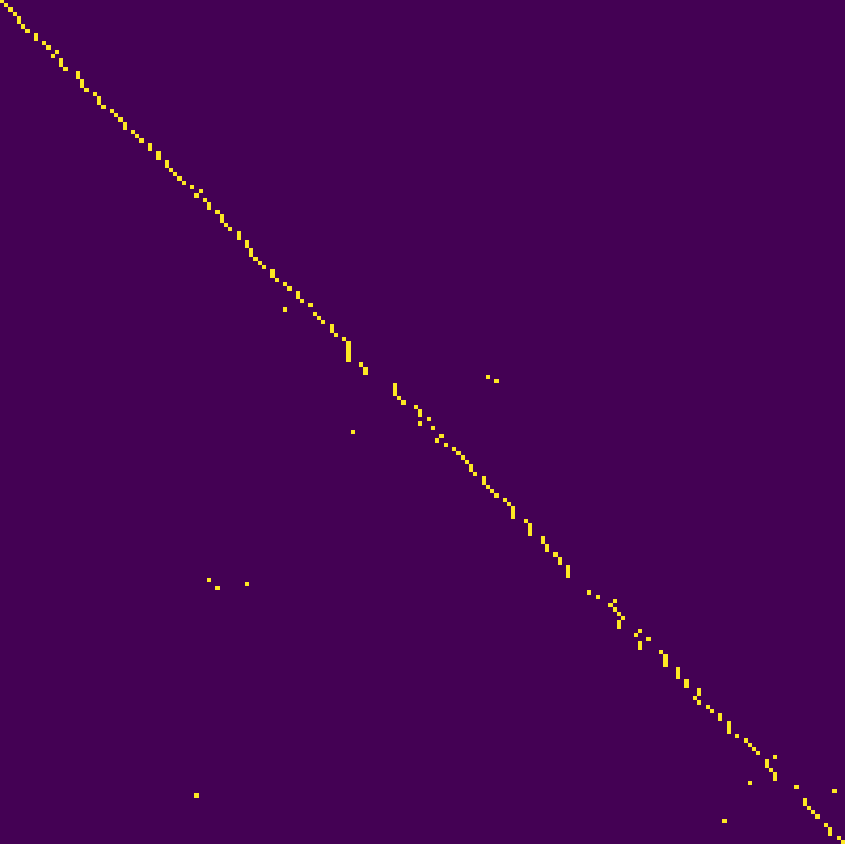}
\hfill
$\mathbf{M}_2$:
\includegraphics[width=0.4\linewidth]{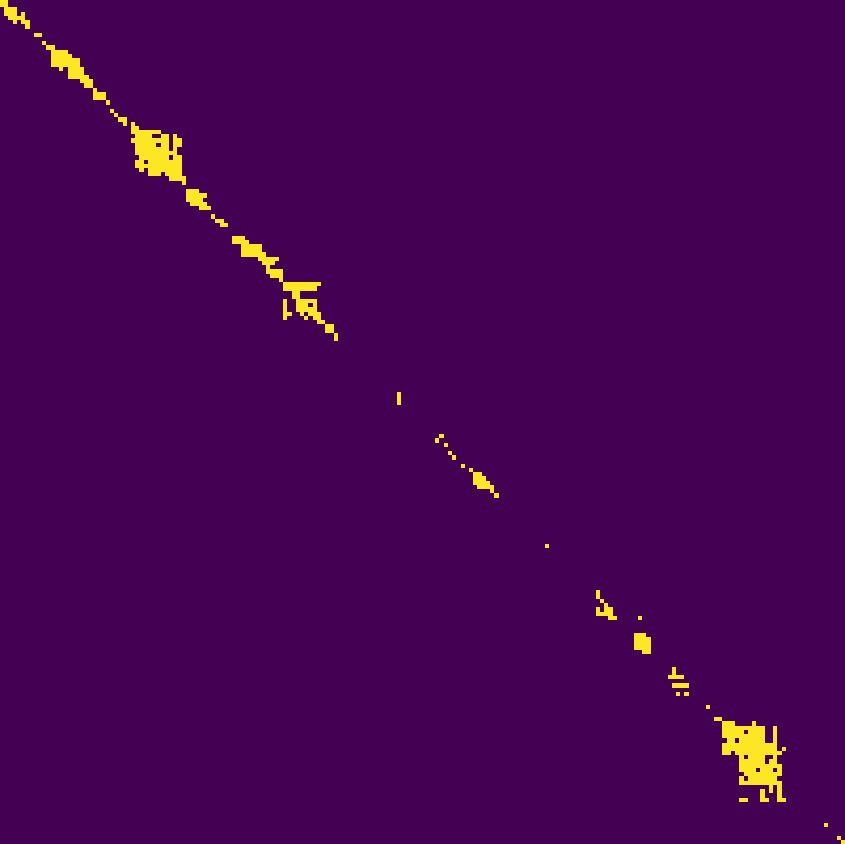}

\vspace{0.1cm}
\raggedright
\noindent Examples for correct and wrong matches from $\mathbf{M}_2$:\hfill

\centering
\includegraphics[width=1\linewidth]{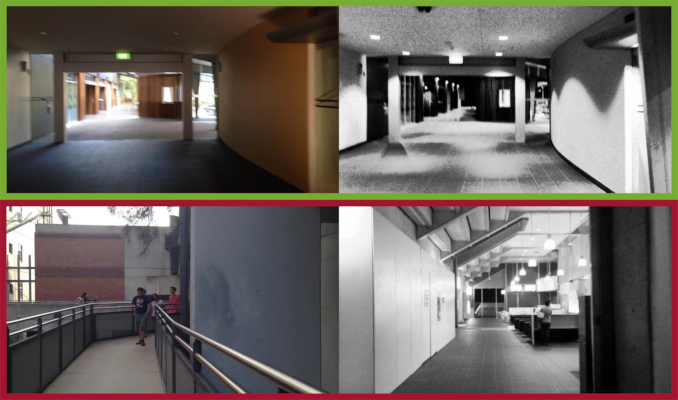}
\end{mdframed}
\caption{The output of a VPR pipeline is typically a set of discrete matchings, i.e.~pairs of query and database images. To obtain matchings for a query image from the similarity matrix (step~4 in \autoref{fig:basic_pipeline}), we can either find the single best matching database image ($\mathbf{M}_1$) or try to find all images in the database that show the same place as the query image ($\mathbf{M}_2$).}
\label{code:finding_matches}
\end{program}

%% file: 5-Evaluation.tex
\section{Evaluation of the performance}
\label{sec:evaluation}

This section is concerned with the evaluation of the matching decisions $\mathbf{M}$, or the pairwise similarities $\mathbf{S}$, which allows the comparison of different VPR methods.
This requires datasets, corresponding ground truth, and performance metrics.
In the following, we outline these components for evaluation and discuss their properties and potential pitfalls.

\subsection{Datasets}\label{sec:datasets}
For VPR, a dataset is composed of one or multiple image sets that have to be matched in order to find shared places.
For example, the popular Nordland dataset \cite{ds_nordland} provides four image sets, one for each season, i.e., spring, summer, fall, and winter. These can be arbitrarily combined for VPR, but a typical choice might be to use \textit{summer} as $DB$ and \textit{spring}, \textit{fall} or \textit{winter} as $Q$.

Existing datasets vary in the type of environment as well as in the type and degree of appearance and viewpoint change:
The \textbf{type of environment} includes indoor environments\ifarxiv~(e.g., Amsterdam-XXXL~\cite{ds_insideout})\fi, urban environments\ifarxiv~(e.g., Oxford RobotCar \cite{ds_robotcar})\fi, suburban environments \ifarxiv (e.g., StLucia Various Times of the Day~\cite{ds_stlucia})~\fi and natural environments like countryside\ifarxiv~(e.g., Nordland \cite{ds_nordland})\fi, forests \ifarxiv(e.g., SFU Mountain \cite{ds_sfu}) \fi or lakes\ifarxiv~(e.g., Symphony Lake \cite{ds_symphony_lake})\fi. 
\textbf{Appearance changes} occur due to 
dynamic objects like pedestrians\ifarxiv~(e.g., GardensPoint Walking \cite{ds_gardenspoint_riverside})\fi,
time of day with lighting changes and moving shadows \ifarxiv (e.g., StLucia Various Times of the Day \cite{ds_stlucia}) \fi
or day versus night\ifarxiv~(e.g., Alderley \cite{seqSLAM})\fi,
weather that is sunny, cloudy, overcast, rainy, foggy, or snowy\ifarxiv~(e.g., Oxford RobotCar \cite{ds_robotcar})\fi,
seasons like spring, summer, fall and winter or dry and wet season\ifarxiv~(e.g., Nordland \cite{ds_nordland})\fi,
elapsed time with roadworks, construction sites or new and demolished buildings up to modern vs.~historical imagery\ifarxiv~(e.g., AmsterTime \cite{ds_amstertime})\fi,
or catastrophic scenarios, e.g., after an earthquake. 
\textbf{Viewpoint changes} between images of the same place range from nearly pixel-aligned \ifarxiv (e.g., Nordland~\cite{ds_nordland}) \fi to left-to-right side of the walkway\ifarxiv~(e.g., GardensPoint Walking~\cite{ds_gardenspoint_riverside})\fi, left-to-right side of the street\ifarxiv~(e.g., SouthBank Bicycle~\cite{ds_gardenspoint_riverside})\fi, panoramic-aligned images to single image\ifarxiv~(e.g., Tokyo 24/7~\cite{Torii15})\fi, panoramic-aligned images to panoramic-aligned images\ifarxiv~(e.g., Pittsburgh250k~\cite{ds_pittsburgh250k})\fi, bikeway-to-street\ifarxiv~(e.g., Berlin Kurf\"urstendamm \cite{ds_berlin})\fi, aerial-to-ground\ifarxiv~(e.g., Danish Airs and Grounds \cite{Vallone2022})\fi, or inside-to-outdoor\ifarxiv~(e.g., Amsterdam-XXXL~\cite{ds_insideout})\fi. For a comprehensive overview of existing datasets, please refer to \cite{Schubert2021,Masone2021}.

\subsection{The ground truth}\label{sec:GT}
Ground truth data tells us which image pairs in a dataset show the same places, and which show different places. This data is necessary for evaluating the results of a place recognition method.
The ground truth is either directly given as a set of tuples indicating which images in the database $DB$ and the query set $Q$ belong to same places, or it is provided via GNSS coordinates or poses using their maximum allowed distances.
Alternatively, some datasets are sampled so that images with the same index in each image set show the same place\ifarxiv~(e.g., Nordland~\cite{ds_nordland} and GardensPoint Walking~\cite{ds_gardenspoint_riverside})\fi.

\subsubsection*{Definition of the ground truth}
To evaluate a VPR result, the definition of a logical ground truth matrix $\mathbf{GT}$ is required. This matrix has the same dimensions as $\mathbf{S}$ and $\mathbf{M}$, i.e., $\mathbf{GT}\in\mathbb{B}^{|Q|\times|Q|}$ or $\mathbf{GT}\in\mathbb{B}^{|DB|\times|Q|}$.
The elements $gt_{ij}\in \mathbf{GT}$ define whether the $i$-th image in $Q$ or $DB$ and the $j$-th image in $Q$ show the same place ($gt_{ij}=true$) or different places ($gt_{ij}=false$).
Their values are set using the ground truth matches from the dataset.

\begin{figure}[t]
    \centering
    \includegraphics{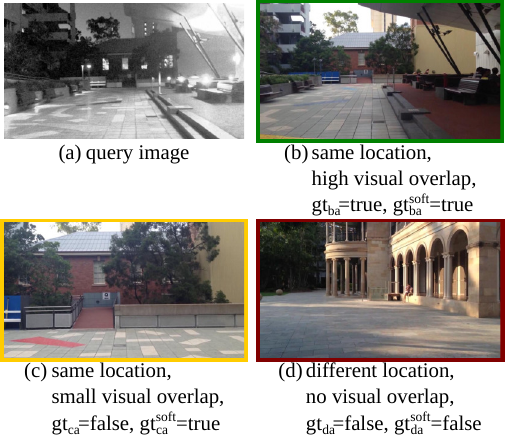}
    \caption{Relation between the visual overlap of a query image (a) and reference images (b-d), and their corresponding ground truth values $gt_{ij}$ and $gt_{ij}^{soft}$. Since (c) and (a) only have a small visual overlap, we do not expect a VPR method to match both images and set $gt_{ca}=false$. However, we also avoid penalization in case the VPR method indeed matches both images by setting $gt_{ca}^{soft}=true$.}
    \label{fig:vis_overlap}
\end{figure}

An additional way of evaluating VPR performance that is used by some researchers is the soft ground truth matrix $\mathbf{GT}^{soft}$. The soft ground truth matrix addresses the problem that we do not expect a VPR method to match images with a very small visual overlap, i.e., $gt_{ij}=false$, as illustrated in~\autoref{fig:vis_overlap}.
However, if a method indeed matches these images with small overlap, we avoid penalization by setting $gt^{soft}_{ij}=true$.
Image pairs without any visual overlap are also labeled $gt^{soft}_{ij}=false$.
Therefore, $\mathbf{GT}^{soft}$ is a dilated version of $\mathbf{GT}$, i.e., it contains all true values contained in $\mathbf{GT}$, as well as additional true values for image pairs with small visual overlap.
Image pairs must be matched if
\begin{equation}
    \mathbf{GT} = true \ .
\end{equation}
Image pairs can, but do not need to be necessarily matched, if
\begin{equation}
    \lnot \mathbf{GT}\land \mathbf{GT}^{soft} = true \ . \label{eq:ignore}
\end{equation}
Note that we use $\lnot$ to denote the logical negation operator. These are usually ignored during evaluation. Image pairs must not be matched if
\begin{equation}
    \mathbf{GT}^{soft} = false \ .
\end{equation}

How $\mathbf{GT}$ and $\mathbf{GT}^{soft}$ are actually used for evaluation is presented in the following.

\begin{figure*}[t]
    \centering
    \includegraphics{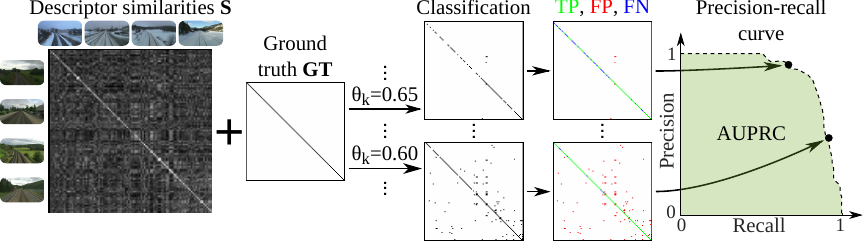}
    \caption{The evaluation pipeline for multi-match VPR, including the precision-recall curve and the area under the precision-recall curve (AUPRC). Given the similarity matrix $\mathbf{S}$ and ground truth $\mathbf{GT}$ and $\mathbf{GT}^{soft}$ (cf.~\autoref{sec:GT}), a range of thresholds $\theta_k\in\{\min({\mathbf{S}}),\dots,\max({\mathbf{S}})\}$ is applied with $\mathbf{S}\geq\theta_k$ to obtain binary matching decisions $m_{ij}\in\mathbf{M}_k$ for each $\theta_k$. In combination with the ground truth, these can be labeled as either True Positives (\textcolor{green}{$TP$}), False Positives (\textcolor{red}{$FP$}), False Negatives (\textcolor{blue}{$FN$}) or True Negatives ($TN$), and converted into a precision-recall curve and the area under the precision-recall curve (AUPRC).}
    \label{fig:AUC}
\end{figure*}

\subsection{Metrics}
This section presents established metrics to evaluate a VPR method, including precision and recall, the precision-recall curve, area under the precision-recall curve, maximum recall at 100\% precision, and recall@$K$~\cite{zaffar2021vpr}. All metrics are implemented in the associated code repository (see~\autoref{sec:intro}).
These metrics are based on
\begin{itemize}
    \item the pairwise descriptor similarities $s_{ij}\in \mathbf{S}$ (cf.~\autoref{sec:S}) or the image matches $m_{ij}\in \mathbf{M}$ (cf.~\autoref{sec:output})
    \item with corresponding ground truth $gt_{ij}\in \mathbf{GT}$ (and $gt^{soft}_{ij}\in \mathbf{GT}^{soft}$ in case the soft ground truth is used).
\end{itemize}

For single-best-match VPR, the evaluation only considers the best matching image pair per query with the highest similarity $s_{i^*j}$:
\begin{equation}
    i^* = \argmax_i{s_{ij}} \ . \label{eq:i_star}
\end{equation}

\subsubsection*{Precision and Recall}
Precision $P$ and recall $R$ are important metrics in the information retrieval domain\ifarxiv~\cite[p.~781]{EncML}\fi.
In the context of VPR, \textbf{precision} $P$ represents the ratio of correctly matched images of same places to the total number of matched images with
\begin{align}
    P = \frac{\#TP}{\#TP+\#FP} \ . \label{eq:P}
\end{align}
\textbf{Recall} $R$ expresses the ratio of correctly matched images of same places to the total number of ground-truth positives~(GTP):
\begin{align}
    R = \frac{\#TP}{\#GTP} \ . \label{eq:R}
\end{align}
In the case of single-best-match VPR, the number of ground-truth positives refers to the total number of query images for which a ground-truth match exists, i.e.,
\begin{equation}
    \#GTP = \sum_{\forall j} \begin{cases}
        1, & \text{if } \exists i: gt_{ij} \\
        0, & \text{otherwise}
    \end{cases} \ ,
\end{equation}
whereas in the case of multi-match VPR, the number of ground-truth positives refers to the number of actually matching image pairs, i.e.,
\begin{equation}
    \#GTP = \sum_{\forall i,j} \begin{cases}
        1, & \text{if } gt_{ij} \\
        0, & \text{otherwise}
    \end{cases} \ .
\end{equation}

$\#TP$ and $\#FP$ are the number of correctly matched and wrongly matched image pairs.
More specifically, \textbf{true positives} TP are actual matching image pairs that were classified as matches:
\begin{equation}
    \#TP = \sum_{\forall i,j}
    \begin{cases}
        1, & \text{if } gt_{ij} \land m_{ij}\\
        0,              & \text{otherwise}
    \end{cases} .
    \label{eq:tp}
\end{equation}
For single-best-match VPR, only $i^*$ from Eq.~\eqref{eq:i_star} is evaluated in Eq.~\eqref{eq:tp} for each query image. The same is true for the following Eq.~\eqref{eq:fp}.

\textbf{False positives} $FP$ are non-matching image pairs that were incorrectly classified as matches:
\begin{equation}
    \#FP = \sum_{\forall i,j}
    \begin{cases}
        1, & \text{if } \lnot gt_{ij} \land m_{ij}\\
        0,              & \text{otherwise}
    \end{cases} .
    \label{eq:fp}
\end{equation}
Note that when using the soft ground truth, image pairs with $\lnot gt_{ij}\land gt^{soft}_{ij} = true$ (cf.~Eq.~\eqref{eq:ignore}) are ignored during the computation of $\#TP$ and $\#FP$.
While \textbf{false negatives} $FN$ are indirectly involved in the calculation of recall $R$, \textbf{true negatives} $TN$ are usually not evaluated due to the typically imbalanced classification problem of VPR with $\#TN\gg \#TP, \#FP, \#FN$. %

\subsubsection*{Precision-recall curve}
Precision-recall curves can be used to avoid actual matching decisions, which are often made after VPR using a computationally expensive verification algorithm.
The idea is to make matching decisions with $\mathbf{M}=\mathbf{S}\geq\theta_k$ over a range of thresholds $\bm{\theta} = \{\min({\mathbf{S}}),\dots,\max({\mathbf{S}})\}$.
For instance, the number of true positives $\#TP_k$ for one specific $\theta_k\in\bm{\theta}$ is then computed with
\begin{equation}
    \#TP_k = \sum_{\forall i,j}
    \begin{cases}
        1, & \text{if } gt_{ij} \land (s_{ij}\geq\theta_k)\\
        0,              & \text{otherwise}
    \end{cases} .
\end{equation}
Following Eqs.~\eqref{eq:P} and \eqref{eq:R}, this leads to two vectors of precision and recall values $P(\theta_k)$ and $R(\theta_k)$, which in combination formulate the \textbf{precision-recall curve}.
The full pipeline for the computation of the precision-recall curve is depicted in \autoref{fig:AUC} and code is provided in \autoref{code:pr_curve}.

\begin{program}[t]
\begin{mdframed}[style=example]
\begin{minted}[fontsize=\footnotesize,escapeinside=||,mathescape=true]{python3}
# precision-recall curve performance evaluation
# for multi-match VPR
from evaluation.metrics import createPR
|$\mathbf{P}$|, |$\mathbf{R}$| = createPR(|$\mathbf{S}$|, |$\mathbf{GT}$|, |$\mathbf{GT}^{soft}$|, matching='multi')
\end{minted}

\par\noindent\rule{\textwidth}{0.4pt}

\noindent Output:

\vspace{0.1cm}

\centering
\noindent
\includegraphics[width=0.9\linewidth]{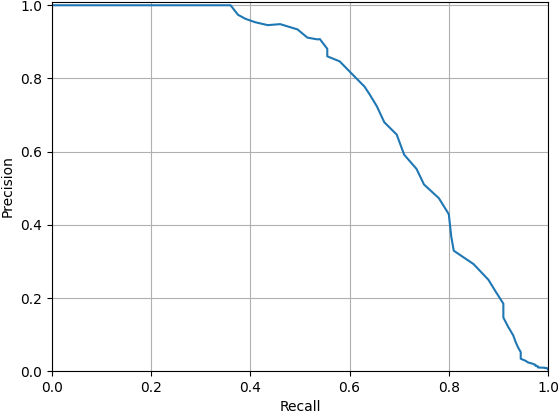}
\end{mdframed}
\caption{To evaluate the quality of a similarity matrix $\mathbf{S}$, we can apply a series of decreasing thresholds $\bm{\theta}$ to match more and more image pairs. Combined with ground-truth information, each threshold leads to a different set of true positives, false positives, true negatives and false negatives, which then provides one point on the precision-recall curve. In this example, we create the precision-recall curve for multi-match VPR.}
\label{code:pr_curve}
\end{program}

\subsubsection*{Area under the precision-recall curve (AUPRC)}
The \textbf{area under the precision-recall curve} (also termed average precision, avgP) can be used to compress a precision-recall curve into a single number, as shown in \autoref{code:auprc}.
In \autoref{fig:AUC}, the AUPRC is visualized as the green area under the precision-recall curve.

\begin{program}[t]
\begin{mdframed}[style=example]
\begin{minted}[fontsize=\footnotesize,escapeinside=||,mathescape=true]{python3}
# evaluate the performance using area under curve
import numpy as np
|$AUPRC$| = np.trapz(|$\mathbf{P}$|, |$\mathbf{R}$|)
\end{minted}

\par\noindent\rule{\textwidth}{0.4pt}

\noindent Output:\\ 
\noindent AUPRC: 0.742
\end{mdframed}
\caption{Finally, to summarize the place recognition quality in a single number, we can use the area under the precision-recall curve (AUPRC). }
\label{code:auprc}
\end{program}

\subsubsection*{Maximum recall at 100\% precision}
The maximum recall at 100\% precision (short \textit{R@100P}) represents the maximum recall where $P=1$ ($100\%$), i.e., the maximum recall without false positives $FP$ (cf.~Eq.~\eqref{eq:fp}).
In the past, this metric was important to evaluate VPR methods for loop closure detection in SLAM.
Keeping the precision at $P=1$ avoids wrong loop closures and, consequently, mapping errors\ifarxiv~\cite{Magnusson2009}\fi.
However, since the advent of robust graph optimization techniques for SLAM \cite{Suenderhauf2012}, the avoidance of wrong loop closures became less relevant.
With robust graph optimization, it is more important to find enough correct loop closures ($TP$) than to avoid wrong loop closures ($FP$). Therefore, using multi-match VPR to identify all loop closures should be preferred over tuning the R@100P for such applications.

If the precision never reaches $P=1$, the maximum recall at 100\% precision is undefined.
Therefore, the maximum recall at 99\% or 95\% precision has been used alternatively.

\subsubsection*{Recall@$K$}
The recall@$K$ (also termed \textit{top-$K$ success rate}) is an often used metric for the evaluation of image classifiers\ifarxiv~\cite[p.~225]{imagenet}\fi.
For place recognition, it is defined as follows: For each query image, given the $K$ database images with the $K$ highest similarities $s_{ij}$, the recall@$K$ measures the rate of query images with at least one actually matching database image. 
That means this metric requires at least one matching image in the database for each query image, which corresponds to a typical localization scenario without exploration.
For mapping with newly visited places, the metric is not defined. In such a scenario, an implementation of recall@$K$ could simply ignore all query images without a matching database image -- however, this workaround would not evaluate the (in)ability of a method to handle exploration during the query run, i.e., new places which are not part of the database set.

The recall@$K$ is particularly suited for visual localization tasks, where the $K$ most similar database or query images are retrieved for a subsequent geometric verification. Note that for VPR in the context of localization without exploration (i.e., all query images have at least one matching reference image), the recall@$1$ and the precision at 100\% recall are identical.

\subsubsection*{Mean, best case and worst case performance}
To get a comprehensive understanding of how well a VPR method performs in different environments, with different types of appearance and viewpoint changes, it is best practice to evaluate it using multiple datasets.
The aforementioned metrics measure the performance on each single dataset. One can get a more condensed view of the overall performance by considering the mean, best case and worst case performance.
The mean performance allows for a quick comparison with other evaluated methods. 
The best case performance shows the maximum achievable performance and reveals potential strengths of an approach:
 if the best case performance of a method is higher than that of the compared methods, this method is well suited for the conditions under which the best case performance was achieved.
The worst case performance reveals the weaknesses of a method and its sensitivity to certain conditions or trajectories (cf.~\autoref{fig:relation_traj_S}). For example, if the worst case performance of a method is lower than the worst case performance of the compared methods, it indicates that this method is less robust and struggles with the specific property of at least one of the evaluated datasets.

We would like to note that there are various other metrics for evaluating VPR methods, including those that take computational time into account. We refer interested readers to~\cite{zaffar2021vpr} for a comprehensive overview, which also includes examples of performance evaluations for the same algorithm across multiple metrics.

%% file: 6-Challenges.tex
\section{Challenges and common ways of addressing them}
\label{sec:challenges}
The previous sections introduced a generic pipeline for VPR, and how to evaluate such a pipeline. In this section, we go beyond this basic pipeline and enlist typical challenges that researchers face in the field of VPR and the ways that prior work has addressed them.

\subsection{Scalability}\label{sec:scalability}
A major challenge in VPR is how to scale up the system to handle large numbers of images in the database or query set. As discussed in \autoref{sec:desc_comp}, holistic image descriptors allow for fast retrieval. %
To reduce the computational effort for descriptor comparison, dimensionality reduction techniques like~\ifarxiv Gaussian, binary, sign or sparse random projection \cite{Dasgupta2000,Achlioptas2003,Li2006}\else random projections\fi, or Principal Component Analysis (PCA) \citemore{}{Liu2012} have been proposed.

However, the computation time for recognizing places is typically still proportional to the number of images in the database $DB$ or query set $Q$.
To further improve efficiency, approximate nearest neighbor search \citemore{}{Li2020} (e.g., a combination of KD-tree~\ifarxiv\cite{kdtree}~\fi and product quantization~\ifarxiv\cite{jegou2010product}~\fi as with DELF~\cite{delf}) can be employed instead of a linear search of all database descriptors, which leads to a sublinear time complexity. %
Additionally, incorporating coarse position data from weak GPS signals can increase efficiency as it reduces the search space~\cite{Vysotska2015}.

Finally, to compensate for the reduced accuracy of holistic descriptors, hierarchical place recognition can be employed. This approach re-ranks the top-$K$ retrieved matches from holistic descriptors through geometric verification with local image descriptors~\citemore{cummins08}{hausler2021patch}.

\subsection{Appearance variations}\label{sec:appearance}
When a robot revisits a place, its current image observation often experiences significant variations in appearance (as discussed in \autoref{sec:datasets}), which can negatively affect the performance.
To reduce the discrepancy between the query observation and the observation stored in the database, techniques such as illumination invariant images~\cite{Maddern2014}, shadow removal~\citemore{mcmanus2014shady}{Corke2013}, appearance change prediction~\citemore{neubert2015superpixel}{Neubert2013}, linear regression~\cite{Lowry2016a}, and deep learning based methods using generative adversarial networks~\citemore{porav2018adversarial}{Anoosheh2019} can be used to convert all images into a reference condition~\cite{Schubert2020}. Such techniques require that the correspondence between each image and its actual condition is provided by human supervision or a condition classifier (e.g., database: summer, query: winter).

To avoid such condition-specific approaches that are trained or designed only for specific conditions (e.g., Night-To-Day-GAN: night and day, shadow removal: different times of day), a condition-wise descriptor standardization can be used to significantly improve performance over a wide range of conditions \cite{Schubert2020}: This standardization normalizes each dimension of the descriptors from one condition to zero mean and unit standard deviation (e.g., once for the database in summer, once for the query set in winter).
Furthermore, if appearance variations occur not only \textit{across} the query and database traverses (e.g., database: sunny, query: rainy) but also \textit{within} a traverse (e.g., database: sunny$\rightarrow$cloudy$\rightarrow$overcast$\rightarrow$rainy, query: sunny), descriptors can be clustered and then standardized per cluster.
Besides addressing individual appearance challenges as above,
a common trend in recent research has been to train deep architectures on large-scale, diverse datasets~\cite{warburg2020mapillary} to achieve global~\cite{Berton_CVPR_2022_cosPlace} and local~\cite{cao2020unifying} descriptors that are robust to appearance variations.
Alternatively, one can combine the strengths of multiple descriptors by simply concatenating them (which sums up their dimensionalities) or combining them using techniques such as hyperdimensional computing (which limits the dimensionality) \cite{Neubert2021}.

\subsection{Viewpoint variations}
A robot may revisit a place from a different viewpoint. For drones, this change could be due to a varying 6-DoF pose, and for an on-road vehicle, it could be due to changes in lanes and direction of travel. In addition to recognizing a local feature or region from different viewpoints, one also needs to deal with often limited visual overlap between an image pair captured from different viewpoints. The problem of viewpoint variations becomes even more challenging when simultaneously affected by appearance variations that widen the scope for perceptual aliasing (the problem of distinct places looking very similar, as discussed in \autoref{sec:intro} and detailed in, e.g.,~\cite{Garg2021}). 
A popular solution to deal with viewpoint variations is to learn  holistic descriptors by aggregating local features in a permutation-invariant manner, that is, independent of their pixel locations, as in NetVLAD~\cite{Arandjelovic_2016_CVPR}.

\subsection{Improving performance}
\label{sec:robustness}
In addition to the approaches mentioned earlier, there are several ways to improve VPR performance by using task-specific knowledge.

\textbf{Sequence-based methods} leverage sequences in the database and query set, which lead to continuous lines of high similarities in the similarity matrix $\mathbf{S}$ (cf.~\autoref{sec:S} and~\autoref{fig:relation_traj_S}).
We can divide these methods into two categories: 
\textbf{Similarity-based sequence methods} use the similarities $s_{ij}\in \mathbf{S}$ to find linear segments of high similarities, e.g., SeqSLAM \cite{seqSLAM}\ifarxiv~or SeqConv \cite{Schubert2021b}\fi, or continuous lines of high similarities with potentially varying slope, e.g., based on a flow network\ifarxiv~\cite{Naseer2014} or on a Hidden Markov Model \cite{Hansen2014}\fi.
\ifarxiv~Methods like SMART \cite{smart} additionally~\else One can also~\fi use available odometry information to find sequences with varying slope.
On the other hand, a sequence of holistic image descriptors can be combined into a single vector. A \textbf{sequence descriptor} defined for a place thus accounts for the visual information observed in the preceding image frames~\cite{SeqNet,mereu2022learning}. These sequence descriptors can be compared between the database and the query sets to obtain place match hypotheses.
\ifarxiv Existing methods include ABLE \cite{Arroyo2015}, MCN \cite{mcn}, delta descriptors \cite{Garg2020}, SeqNet \cite{SeqNet}, and different deep learning based approaches \cite{Facil2019,mereu2022learning}.\fi

Besides leveraging descriptor similarities $\mathbf{S}$ \textit{between} the database and query sets, \textbf{intra-database and intra-query similarities $\mathbf{S}^{DB}$ and $\mathbf{S}^Q$}, i.e., descriptor similarities \textit{within} the database and query sets, can be used to improve performance. For example, in~\cite{Schubert2021b}, the intra-set similarities $\mathbf{S}^{DB}$ and $\mathbf{S}^Q$ are used in combination with $\mathbf{S}$ and sequence information to formulate a factor graph that can be optimized to refine the similarities in $\mathbf{S}$. In this graph, the intra-set similarities are used to connect images within the database or query sets that are likely to show the same or different places due to a high or low intra-set similarity. For example, let us suppose that the $l$-th query image has high similarities $s_{il}$ and $s_{jl}$ to the $i$-th and $j$-th database images. Let us further suppose that the similarity $s_{kl}$ to the $k$-th database image is low, although the $i$-th, $j$-th and $k$-th database images have high intra-database similarities $s_{ij}^{DB}$, $s_{ik}^{DB}$ and $s_{jk}^{DB}$. The graph optimization then detects that the similarity $s_{kl}$ between the $k$-th database image and the $l$-th query image is also likely to be high.

Methods such as experience maps \cite{Churchill2013} and co-occurrence maps \cite{Johns2013} can be used in cases where the robot \textbf{frequently revisits the same places}.
These ``memory-based'' methods continually observe each place and create a descriptor every time the appearance changes.
During a comparison of a new query descriptor with this aggregated ``map'', only one descriptor of a similar condition needs to be matched to recognize the place, reducing the need for condition-invariant descriptors\ifarxiv~\cite{molloy2020intelligent}\fi. Several approaches go beyond these memory-based techniques by modeling spatiotemporal dynamics to forecast feature persistence~\cite{rosen2016towards}, expected outdoor conditions~\cite{linegar2015work}, or map occupancy~\cite{krajnik2017fremen}.

In the case of robot localization with known places (i.e., each visited query place is guaranteed to be in the database and no exploration beyond this mapped area is performed), VPR can benefit from \textbf{place-specific classifiers}, which can improve accuracy with reduced map storage or retrieval time~\citemore{mcmanus2015learning}{gronat2013learning,McManus2014,keetha2021hierarchical}.
A similar approach is to train a deep learning-based \textbf{place classifier} that directly outputs a place label for a given image\ifarxiv~\cite{Weyand2016,HussainiRAL2022ICRA2022}\fi, or to create environment-specific descriptors\ifarxiv~\cite{Neubert22}\fi. %
Another direction is to exploit known place types for place type matching to limit the number of potential matches between the database and query set \cite{sunderhauf2015performance}. For example, instead of searching through all database images, if the query image was taken in a forest, such semantic categorization constrains the database images to only those that were also taken in a forest.

%% file: 7-Conclusions.tex
\section{Conclusions}
\label{sec:conclusions}
Visual Place Recognition (VPR) is a well established problem that has found widespread interest and use in both computer vision and robotics. In this tutorial, we have described the visual place recognition task, including its various problem categories and subtypes, their typical use cases, and how it is typically implemented and evaluated. Additionally, we discussed a number of methods that can be used to address common challenges in VPR.

There are a number of open challenges such as system integration, enriched reference maps, view synthesis, and the design of a ``one-fits-all'' solution that still need to be tackled by the community. While we do not discuss these challenges in this tutorial, we refer the interested reader to~\cite{Schubert2021,Garg2021}.